\documentclass[letterpaper, 10 pt, conference]{ieeeconf}
\IEEEoverridecommandlockouts{}

\newcommand\copyrightnotice{%
\begin{tikzpicture}[remember picture,overlay]
\node[anchor=south,xshift=5pt,yshift=10pt] at (current page.south) {%
\begin{minipage}{0.95\textwidth}
\centering
\footnotesize
\copyright~2026 IEEE. Personal use of this material is permitted.
Permission from IEEE must be obtained for all other uses, in any current
or future media, including reprinting/republishing this material for
advertising or promotional purposes, creating new collective works,
for resale or redistribution to servers or lists, or reuse of any
copyrighted component of this work in other works.
\end{minipage}
};
\end{tikzpicture}%
}

\usepackage{etoolbox} 
\usepackage{graphicx}
\usepackage[table]{xcolor}
\usepackage[utf8]{inputenc}
\usepackage{mathptmx}
\usepackage{array,contour}
\usepackage{tabularx}
\usepackage{booktabs}
\usepackage{multirow}
\usepackage{ragged2e}
\usepackage[T1]{fontenc}

\usepackage{parskip}
\usepackage{amsmath, amssymb, amsbsy, bm}
\usepackage{amsfonts}
\usepackage{mathtools}
\usepackage{glossaries}

\usepackage{float}
\usepackage{siunitx}

\usepackage{tikz, pgfplots}
\usetikzlibrary{spy}
\usetikzlibrary{backgrounds}
\usetikzlibrary{decorations}
\usetikzlibrary{shapes,snakes}
\usetikzlibrary{decorations.markings}
\usepgfplotslibrary{groupplots}
\usetikzlibrary{shapes.geometric}
\usetikzlibrary{calc}
\usetikzlibrary{positioning}
\usetikzlibrary{fit}
\usetikzlibrary{matrix}
\usepackage{tikzscale}

\usepackage{grffile}
\usepackage{currfile}
\usepackage{caption}
\usepackage{subcaption}

\usepackage[english]{babel}
\usepackage{csquotes}
\usepackage{pgfplots}

\usepgfplotslibrary{groupplots,dateplot}
\usetikzlibrary{positioning,arrows.meta}   
\usepgflibrary{patterns}
\pgfplotsset{compat=newest}
\usepackage[backend=biber,style=ieee,maxbibnames=3]{biblatex}

\makeglossaries
\usepackage[mathcal]{euscript}

\usepackage{placeins}
\usepackage{censor}

\def\BibTeX{{\rm B\kern-.05em{\sc i\kern-.025em b}\kern-.08em
    T\kern-.1667em\lower.7ex\hbox{E}\kern-.125emX}}

\newtoggle{todos}
\settoggle{todos}{false}
\newif\ifdraft\draftfalse{}

\usepackage{xargs} 
\usepackage{marginnote}
\usepackage[colorinlistoftodos,textsize=small]{todonotes}

\makeatletter
\newcommand\@todonotes@owner{default}
\define@key{todonotes}%
    {owner}{\def\@todonotes@owner{#1}}

\newtoggle{ownerdefault}
\newtoggle{ownerB}

\newcommand\marginnoteinfigure[1]{
\ifnum\@floatpenalty<0\relax{\marginnote{#1}}\else{#1}\fi
}
\newcommand\inlineinfigure{
\ifnum\@floatpenalty<0\relax{inline}\else{noinline}\fi
}
\newcommand\iffigure[2]{
\ifnum\@floatpenalty<0\relax #1 \else #2 \fi
}
\newcommand\isinfigure{
\ifnum\@floatpenalty<0\relax{true}\else{false}\fi
}

\newcommand\todocap{}
\newcommand\todotext{}

\newcommandx\todonewText[3][1,3]{
    \iftoggle{todos}{\todo[%
    owner=default,
    caption={\ifstrempty{#3}{}{\Issue{#3} }\ifstrempty{#1}{#2}{#1}},color=red!20%
    ]%
    {#2\ifstrempty{#3}{}{ \Issue{#3}}}
    }{}}
\newcommandx\todonewFig[3][1,3]{
    \iftoggle{todos}{\marginnote{\todo[%
    owner=default,inline,
    caption={\ifstrempty{#3}{}{\Issue{#3} }Fig: {\ifstrempty{#1}{#2}{#1}}},color=red!20%
    ]%
    {Fig: {#2}\ifstrempty{#3}{}{ \Issue{#3}}}}
    }{}}
  \newcommandx\todonew[3][1,3]{
  \renewcommand\todocap{\ifstrempty{#3}{}{\Issue{#3} }\iffigure{Fig: }{}{\ifstrempty{#1}{#2}{#1}}}%
  \renewcommand\todotext{\iffigure{Fig: }{}{#2}\ifstrempty{#3}{}{ \Issue{#3}}}%
  \iftoggle{todos}{\marginnoteinfigure{%
  \iffigure{\todo[owner=default,caption={\todocap},color=red!20,inline]%
  {\todotext}%
  }{%
    \todo[owner=default,caption={\todocap},color=red!20]%
  {\todotext}%
  }}
  }{}}
\newcommandx\tododoneText[3][1,3]{
    \iftoggle{todos}{\todo[%
    owner=B,
    caption={\ifstrempty{#3}{}{\Issue{#3} }\ifstrempty{#1}{#2}{#1}},color=green!50!white%
    ]%
    {#2\ifstrempty{#3}{}{ \Issue{#3}}}
    }{}}
\newcommandx\tododoneFig[3][1,3]{
    \iftoggle{todos}{\marginnote{\todo[%
    owner=B,inline,
    caption={\ifstrempty{#3}{}{\Issue{#3} }Fig: {\ifstrempty{#1}{#2}{#1}}},color=green!50!white%
    ]%
    {Fig: {#2}\ifstrempty{#3}{}{ \Issue{#3}}}}
    }{}}
  \newcommandx\tododone[3][1,3]{
  \renewcommand\todocap{\ifstrempty{#3}{}{\Issue{#3} }\iffigure{Fig:}{} {\ifstrempty{#1}{#2}{#1}}}%
  \renewcommand\todotext{\iffigure{Fig:}{} {#2}\ifstrempty{#3}{}{ \Issue{#3}}}%
  \iftoggle{todos}{\marginnoteinfigure{%
  \iffigure{\todo[owner=B,caption={\todocap},color=green!50!white,inline]%
  {\todotext}%
  }{%
    \todo[owner=B,caption={\todocap},color=green!50!white]%
  {\todotext}%
  }}
  }{}}

\ifdraft
\renewcommand{\@todonotes@addElementToListOfTodos}{%
    \if@todonotes@colorinlistoftodos%
     \addtocontents{tdo}
      {%
       \protect\iftoggle{owner\@todonotes@owner}
         {%
          \protect\contentsline {todo}
            {\protect\fcolorbox{\@todonotes@currentbordercolor}%
                {\@todonotes@currentbackgroundcolor}%
                {\protect\textcolor{\@todonotes@currentbackgroundcolor}{o}}%
            \ \@todonotes@caption
            }{\thepage}{\@currentHref}%
          }{}%
       }%
    \else%
      \addtocontents{tdo}
      {%
       \protect\iftoggle{owner\@todonotes@owner}
         {%
          \protect\contentsline {todo}
            {\@todonotes@caption
            }{\thepage}{\@currentHref}%
          }{}%
       }%
    \fi}%
\fi

\makeatother

\iftoggle{todos}{
\paperwidth=\dimexpr \paperwidth+9cm\relax
\oddsidemargin=\dimexpr \oddsidemargin+4.5cm\relax
\evensidemargin=\dimexpr \evensidemargin+4.5cm\relax
\marginparwidth=\dimexpr \marginparwidth + 3cm\relax
\hoffset=\dimexpr \hoffset\relax
}
{}

\usepackage{booktabs}
\usepackage{siunitx}
\usepackage{multirow}
\usepackage{graphicx}

\begin{document}
\graphicspath{{figures/}}

\title{CANMOT:\ Class-Aware Noise Modeling for Multi-Object Tracking in Autonomous Driving}

\author{Timo Osterburg, Stefan Schütte, and Torsten Bertram\\
\small Institute of Control Theory and Systems Engineering, TU Dortmund University, Germany.
}

\maketitle

\copyrightnotice

\begin{abstract}
Kalman filter (KF)-based multi-object tracking (MOT) remains a strong baseline for autonomous driving due to its strong performance, computational efficiency and interpretability.
In most practical systems, the process noise and measurement noise covariances are defined globally and shared across object classes, presuming identical uncertainty characteristics across heterogeneous traffic participants.

This work revisits this assumption and proposes CANMOT, a class-aware and object-aligned noise modeling framework for KF-based 3D MOT.
Class-specific diagonal process and measurement covariance matrices are introduced and optionally expressed in the object coordinate frame to preserve longitudinal-lateral anisotropy.

Systematic experiments on the nuScenes benchmark show that class-aware and object-aligned noise modeling improves tracking performance and substantially reduces identity switches compared to state-of-the-art (SotA).
In addition, the consistency of the estimated uncertainty is analyzed using the Average Normalized Estimation Error Squared (ANEES) and \(\chi^2\)-based violation tests.
The results reveal severe overconfidence in standard KF-based MOT baselines.
While the proposed formulation improves calibration without modifying the underlying filtering framework, it still exhibits substantial inconsistency, highlighting the need for further research in this area.

Code is available at \url{https://github.com/rst-tu-dortmund/CANMOT}.

\end{abstract}


\newacronym{amota}{AMOTA}{Average Multi Object Tracking Accuracy}
\newacronym{amotp}{AMOTP}{Average Multi Object Tracking Precision}
\newacronym{anees}{ANEES}{Average Normalized Estimation Error Squared}
\newacronym{bev}{BEV}{bird-eye-view}
\newacronym{cv}{CV}{constant-velocity}
\newacronym{fns}{FN}{False Negatives}
\newacronym{fps}{FP}{False Positives}
\newacronym{frag}{FRAG}{Fragmentation}
\newacronym{ids}{IDS}{ID switch}
\newacronym{iou}{IoU}{intersection-over-union}
\newacronym{kf}{KF}{Kalman filter}
\newacronym{map}{mAP}{mean Average Precision}
\newacronym{ml}{ML}{Machine Learning}
\newacronym{mot}{MOT}{multi-object tracking}
\newacronym{nees}{NEES}{Normalized Estimation Error Squared}
\newacronym{nds}{NDS}{nuScenes Detection Score}
\newacronym{nms}{NMS}{non-maximum suppression}
\newacronym{sota}{SotA}{State of the Art}
\newacronym{rois}{RoIs}{regions of interest}
\newacronym{tps}{TP}{True Positives}
\newacronym{gt}{GT}{Ground-Truth}

\section{Introduction}
\label{sec:intro}

\Gls{mot} is a central component of autonomous driving systems.  
It provides temporally consistent object states that are required for prediction, planning, and decision making.

Many real-time \gls{mot} pipelines combine a 3D object detector with a motion prediction module inside a \gls{kf} framework.  
The detector provides noisy per-frame measurements with covariance $\bm{R}$, while the motion model propagates object states subject to process noise covariance \(\bm{Q}\).  
The \gls{kf} estimates the current state based on uncertain prediction and measurement.
Here, the parameterization of \(\bm{Q}\) and \(\bm{R}\) fundamentally dictates estimation accuracy.

\begin{figure}[t]
    \centering
    \includegraphics[width=0.48\textwidth]{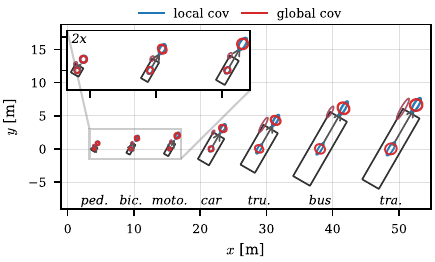}
    \caption{
        Comparative overview of sample covariance estimation depicted by ellipses in the local object (blue) and global map (red) coordinate frames for each individual object class on the nuScenes dataset.
        Classes from left to right: pedestrian, bicycle, motorcycle, car, truck, bus, trailer.
        }
    \label{fig:covariance_overview}
\end{figure}

Figure~\ref{fig:covariance_overview} illustrates the sample covariance computed on nuScenes~\cite{nuscenes} trainval of the measurement and process noise for a constant velocity model of different object classes in both the local object and global map coordinate frames.
The measurement sample covariance for the center position is shown as ellipses around the center.
The variance in extent is shown as ellipses around the front left corner of the bounding box.
The covariance of the center predicted with constant velocity under process noise is shown as ellipses around the predicted center.

As illustrated in Figure~\ref{fig:covariance_overview}, object classes exhibit distinct perceptual characteristics that influence measurement uncertainty.  
Smaller objects such as pedestrians or motorcycles generate fewer LiDAR returns, which increases localization variance.  
Elongated vehicles are often only partially observed, leading to ambiguity in extent and center estimation.  
Compact or near-symmetric objects yield less reliable heading estimates than long vehicles with clearly defined principal axes.  
Velocity estimation from multi-sweep accumulation is inherently more sensitive to shape irregularities and partial observations than for rigid vehicle contours.
These effects imply class-dependent measurement uncertainty.

Figure~\ref{fig:covariance_overview} shows that different object categories also exhibit distinct motion characteristics.  
Vehicles obey steering geometry that restricts feasible lateral motion and induces structured velocity evolution.  
Pedestrians are not subject to such constraints and may change direction abruptly.  
Motorcycles and light vehicles can exhibit higher acceleration and jerk than heavy trucks or buses, resulting in larger deviations from constant-velocity assumptions.  
These differences imply class-dependent motion model uncertainty.

A common simplifying assumption in practical \gls{kf}-based \gls{mot} systems is to use one $\bm{Q}$ and $\bm{R}$ defined in the map coordinate frame.  
This assumes similar uncertainty characteristics across all object classes and directions.
The work at hand models both $\bm{Q}$ and $\bm{R}$ in a class-aware manner, yielding class-specific noise parameters $(\bm{Q}_c, \bm{R}_c)$.

Furthermore, uncertainty in localization and motion is inherently directional.  
Longitudinal and lateral motion exhibit different variability due to physical constraints.  
When noise is defined in global coordinates, these directional effects are averaged over varying object orientations, resulting in effectively isotropic modeling.  
Therefore, we propose to express noise in the individual object coordinate frame to preserve longitudinal-lateral anisotropy.

Accurate uncertainty modeling is critical for downstream planning modules that explicitly incorporate state covariance.
Consistent uncertainty estimates are required to reliably and safely react to the environment.
While overestimated uncertainty may lead to unnecessary conservatism, underestimated uncertainty may lead to unsafe behavior.
Therefore, it is crucial to investigate the consistency of the estimated uncertainty with the true error of the filter.
To the best of the authors' knowledge, this aspect has not been systematically evaluated in \gls{kf}-based 3D \gls{mot} benchmarks for autonomous driving. 

The proposed formulation is evaluated on the nuScenes benchmark~\cite{nuscenes}.  
Systematic experiments isolate the impact of design choices under a constant-velocity motion model.

To model class-dependent and anisotropic uncertainty in \gls{kf}-based \gls{mot}, we propose CANMOT, a novel approach that optimizes class-aware noise parameters in an object-aligned coordinate frame.
The work at hand addresses the following research questions:
\begin{itemize}
    \item R1: Does modeling noise in an object-aligned coordinate frame improve tracking performance compared to a global coordinate frame?
    \item R2: Does class-aware noise modeling improve tracking performance compared to shared noise parameters?
    \item R3: How do the proposed design choices affect the consistency of the estimated uncertainty with the true error?
\end{itemize}

\tododone{NEES/Consistency in Motivation}
\section{Related Work}
\label{sec:related_work}
\todonew{Related work mit Weak in the NEES, NEES im MOT-Kontext, genereller Wahl der Parameter für KF in MOT}
To situate the work at hand within the field of \gls{mot}, we discuss the differences in optimization strategies employed by \gls{sota} methods.

On one end, methods like \cite{Zaech2022} 
aim to learn the entire tracking task purely from data.
This approach can handle any class-specific variation in detection quality implicitly, but its consistency is difficult to assess due to the black box nature.

AB3DMOT \cite{AB3DMOT} uses a \gls{kf} for \gls{mot} in 3D, but chooses noise covariance matrices shared for all classes seemingly arbitrarily in global coordinates.
Here, we see optimization potential with careful choice of the parameters.

Other approaches aim to calculate a motion process covariance matrix from data \cite{chiu2020probabilistic3dmultiobjecttracking}.
Here, acceleration characteristics are estimated from data and expanded to a diagonal covariance for acceleration and velocity.
We compute the full sample covariances and optimize the noise covariance matrices with respect to the AMOTA metric.

\citeauthor{Fleck2021} \cite{Fleck2021} investigate Bayesian optimization for MOT with simulated data.
Here, scalars are optimized for $\bm{Q}$ and $\bm{R}$ and knowledge of the motion process is used to derive factors based on $\Delta t$ to expand them to full covariance matrices.

Tuning covariance matrices of a \gls{kf} via Bayesian optimization to achieve an improvement in certain metrics is a well known technique in state estimation \cite{Chen2018weaknees}.
However, to our knowledge, no thorough investigation of its application to \gls{mot} has yet been conducted.

Recently, \cite{DiffKFCAV2024} proposed estimating covariance matrices from sensory data using \gls{ml} techniques for \gls{mot}.
However, the approach operates in global coordinates and only estimates the measurement covariance.
The motion model noise covariance matrix is left as a diagonal matrix.

\citeauthor{Cho2014} \cite{Cho2014} take a polyhedral approach to MOT, choosing different motion models per object class.
$\bm{R}$ is described to be derived from data, while $\bm{Q}$ is not explicitly described.
The approach uses the ego-vehicle frame for computations related to covariances.

Poly-MOT~\cite{poly_mot} was a long-term leader of the nuScenes \cite{nuscenes} tracking challenge benchmark.
The authors propose use of different motion models per class and separately tune $\bm{Q}$ and $\bm{R}$ per motion model by hand.

IMM-MOT~\cite{IMM_MOT} switches between multiple motion models based on likelihood derived from residuals.
It is built upon Poly-MOT~\cite{poly_mot} and retains the same covariance matrices.
These advances are perpendicular to our approach, and could be combined in future work.

NeMOT~\cite{nemot} learns the prediction and the measurement function.
Using particles within a factor graph, the approach implicitly estimates the covariance of the state estimate by the distribution of particles.

MCTrack~\cite{MCTrack} performs a decoupled estimation of position, size and heading angle in the \gls{mot} context.
This, together with a new association metric forms their main contribution.
Investigation of the provided code release shows that class-wise hand-tuned covariance matrices are also employed.
The choice of class-wise covariance matrices remains unaddressed and lacks empirical justification in the paper.

\section{Problem Formulation}
\label{sec:problem}

Given a sequence of detections over time, the objective of \gls{mot} is to estimate temporally consistent object states and identities under motion and measurement uncertainty.

Each tracked object is represented by the state vector
\[
\bm{x} =
\begin{bmatrix}
x & y & z & w & l & h & v_x & v_y & v_z & \theta
\end{bmatrix}^\top,
\]
where $(x,y,z)$ denote the object center in the global coordinate frame, $(w,l,h)$ its spatial extent, $(v_x,v_y,v_z)$ its velocity components, and $\theta$ its yaw angle.

State propagation is performed using a linear \gls{cv} model within a \gls{kf}.
The state evolves according to
\[
\bm{x}_{k+1} = \bm{F}\bm{x}_k + \bm{w}_k,
\qquad
\bm{w}_k \sim \mathcal{N}(\bm{0}, \bm{Q}),
\]
where $\bm{F}$ denotes the transition matrix, $\bm{Q}$ the process noise covariance matrix, and $k$ the discrete time index.
The transition matrix propagates positions using velocities while keeping extent and orientation constant.
Let $\bm{P}_k$ denote the covariance matrix of the state estimate at time step $k$.
The superscript $(\cdot)^-$ denotes a priori quantities, while $(\cdot)^+$ denotes a posteriori quantities.
Quantities denoted with \(\widehat{(\cdot)}\) are filter estimates. 
The predicted state and covariance are
\[
\widehat{\bm{x}}^-_{k+1} = \bm{F}\widehat{\bm{x}}^+_k,
\qquad
\widehat{\bm{P}}^-_{k+1} = \bm{F}\widehat{\bm{P}}^+_k\bm{F}^\top + \bm{Q}.
\]

The detector provides measurements
\[
\bm{z} =
\begin{bmatrix}
x & y & z & w & l & h & v_x & v_y & \theta
\end{bmatrix}^\top,
\]
which relate linearly to the state through
\[
\bm{z}_k = \bm{H}\bm{x}_k + \bm{v}_k,
\qquad
\bm{v}_k \sim \mathcal{N}(\bm{0}, \bm{R}),
\]
where $\bm{H}$ denotes the linear measurement projection matrix and $\bm{R}$ the measurement noise covariance matrix.
Thus, a natural choice for the initial state covariance of the initial state \(\bm{x}_0 = \bm{H}^\top\bm{z}_0\) is \(\bm{P}_0 = \bm{H}^\top\bm{R}\bm{H}\).
As $v_z$ is not measured, a constant is set for the corresponding diagonal entry in $\bm{P}_0$.

The tracking pipeline follows the two-stage class-aware matching paradigm introduced in Poly-MOT~\cite{poly_mot}.
Association is performed using a geometric similarity measure based on 3D GIoU.
Optimal assignment is computed via the Hungarian algorithm.
Matched tracks are updated using the \gls{kf}.
Unmatched detections spawn new tracklets if they pass a score threshold.
Unmatched tracks undergo an exponential decay in score and are retained if their score remains above a predefined threshold.
They are removed after a fixed number of consecutive misses.

For matched track-detection pairs, the linear \gls{kf}~\cite{kalman1960} update is performed as
\begin{align*}
    \bm{K}_k &= \widehat{\bm{P}}^-_k \bm{H}^\top
(\bm{H}\widehat{\bm{P}}^-_k\bm{H}^\top + \bm{R})^{-1},\\
\widehat{\bm{x}}_k^+ &= \widehat{\bm{x}}^-_k + \bm{K}_k(\bm{z}_k - \widehat{\bm{z}}_k),\\
\widehat{\bm{P}}_k^+ &= (\bm{I} - \bm{K}_k\bm{H})\widehat{\bm{P}}^-_k,
\end{align*}
where $\widehat{\bm{z}}_k = \bm{H}\widehat{\bm{x}}^-_k$, $\bm{K}_k$ denotes the Kalman gain, and $\bm{I}$ is the identity matrix.
Angular residuals for the yaw component are wrapped to the interval $(-\pi,\pi]$ prior to the update.

In typical \gls{kf}-based \gls{mot} systems, the process noise and measurement noise are defined as diagonal and shared across all object classes.

Common autonomous driving benchmarks, such as nuScenes~\cite{nuscenes} and Waymo Open Dataset~\cite{waymo}, provide annotations and evaluation in global map coordinates.
Consequently, uncertainty modeling is typically performed in this frame.
This global formulation does not explicitly account for class-dependent or orientation-dependent uncertainty characteristics.
\section{Method}
\label{sec:method}
\begin{table*}[t]
\centering
\caption{Comparison of tracking approaches on nuScenes val using standard \gls{mot} metrics alongside the mean \gls{anees}.
The best results are highlighted in \textbf{bold}. Second best results are underlined.}
\label{tab:tracking_comparison}

\setlength{\tabcolsep}{3pt}
\renewcommand{\arraystretch}{1.15}

\resizebox{\textwidth}{!}{%
\begin{tabular}{
l
c c c c c c c c
c
c
c
c
c
c c c c
c
}

\toprule

& \multicolumn{8}{c}{AMOTA $\uparrow$}
& {AMOTP $\downarrow$}
& {IDS $\downarrow$}
& {FRAG $\downarrow$}
& {TP $\uparrow$}
& {FP $\downarrow$}
& \multicolumn{4}{c}{TP Errors $\downarrow$}
& {mANEES}
\\

\cmidrule(lr){2-9}
\cmidrule(lr){15-18}

Approach
& all
& Bic.
& Bus
& Car
& Motor.
& Ped.
& Tra.
& Tru.
&
&
&
&
&
& $e_\mathrm{t}$
& $e_\mathrm{s}$
& $e_\mathrm{v}$
& $e_\mathrm{o}$
&
\\

\midrule

\multicolumn{19}{c}{\cellcolor{gray!20}\textbf{Local}} \\
\midrule

CANMOT-SC
& 73.5
& \(\bm{56.7}\)
& \(\bm{87.6}\)
& \(\bm{86.5}\)
& 78.1
& \(\bm{83.4}\)
& 50.9
& \(\underline{71.3}\)
& \(\underline{51.2}\)
& \(\bm{190}\)
& 285
& 83876
& \(\bm{12251}\)
& \(\bm{.29}\)
& \(\bm{.20}\)
& \(\bm{.26}\)
& .28
& \(\bm{20.4}\)
\\

CANMOT-Opt
& \(\underline{73.8}\) & \(\bm{56.7}\) & \(\underline{87.5}\) & \(\bm{86.5}\) & \(\underline{78.2}\) & \(\underline{83.3}\) & \(\bm{53.0}\) & 71.1
& 52.2
& 208 & 285 & 84340 & 12957
& \(\underline{.30}\) & \(\underline{.21}\) & .28 & .27
& 2092.3
\\

Shared-Opt
& 73.3 & 56.2 & 86.9 & 86.2 & 77.7 & 83.1 & \(\underline{51.6}\) & \(\underline{71.3}\)
& 52.2
& 226 & 316 & 84550 & 13413
& \(\underline{.30}\) & \(\underline{.21}\) & \(\bm{.26}\) & \(\bm{.25}\)
& 1902.5
\\

SC-T
& 72.0
& 53.7
& 84.6
& 86.2
& 76.6
& 82.8
& 49.8
& 70.4
& 55.8
& 246
& 273
& 84547
& 13563
& .33
& \(\bm{.20}\)
& \(\bm{.26}\)
& .28
& 52.3
\\

SC-V
& 72.0
& 53.8
& 84.7
& 86.2
& 76.9
& 82.9
& 49.2
& 70.1
& 55.4
& 221
& 269
& 83857
& 12959
& .33
& \(\bm{.20}\)
& \(\bm{.26}\)
& .29
& 43.3
\\

SC-TV
& 71.9
& 53.7
& 84.6
& 86.2
& 76.6
& 82.9
& 49.3
& 70.4
& 55.8
& 240
& 271
& 84372
& 13381
& .33
& \(\bm{.20}\)
& \(\bm{.26}\)
& .28
& 49.4
\\

\midrule
\multicolumn{19}{c}{\cellcolor{gray!20}\textbf{Global}} \\
\midrule


Shared-Opt
& 73.0 & 55.4 & 87.4 & 86.3 & 76.8 & 83.0 & 51.2 & 71.0
& \(\bm{50.9}\)
& \(\underline{204}\) & 289 & 83474 & \(\underline{12442}\)
& \(\bm{.29}\) & \(\bm{.20}\) & \(\underline{.27}\) & \(\underline{.26}\)
& 7545.1
\\

SC-T
& 72.1
& 54.9
& 84.9
& 86.1
& 76.0
& 83.0
& 49.6
& 70.5
& 54.7
& 252
& \(\underline{263}\)
& 83971
& 13030
& .33
& \(\bm{.20}\)
& \(\underline{.27}\)
& .29
& 51.8
\\

SC-V
& 71.6
& 52.7
& 84.8
& 86.1
& 75.7
& 82.8
& 49.0
& 70.2
& 55.2
& 271
& 264
& 84307
& 13533
& .33
& \(\bm{.20}\)
& \(\underline{.27}\)
& .29
& \(\underline{42.3}\)
\\

SC-TV
& 72.1
& 54.9
& 84.9
& 86.1
& 75.9
& 83.1
& 49.5
& 70.4
& 54.8
& 248
& \(\bm{249}\)
& 83808
& 12840
& .33
& \(\bm{.20}\)
& \(\underline{.27}\)
& .30
& 48.7
\\

\midrule
\multicolumn{19}{c}{\cellcolor{gray!20}\textbf{\glsentrylong{sota} (global)}} \\
\midrule

Poly-MOT\(^{\star\diamond}\)~\cite{poly_mot}
& 72.8 & 54.4 & 87.3 & 86.3 & 76.8 & 83.0 & 51.0 & 71.2
& 51.7
& 235 & 295 & 84094 & 13062
& \(\bm{.29}\) & \(\underline{.21}\) & 1.25 & \(\bm{.25}\)
& $6\times10^{14}$
\\

MCTrack\(^{\star}\)~\cite{MCTrack}
& \(\bm{74.0}\) & -- & -- & -- & -- & -- & -- & --
& --
& 275 & -- & \(\underline{85900}\) & 13083
& -- & -- & -- & --
& --
\\

IMM-MOT\(^{\star\ddagger}\)~\cite{IMM_MOT}
& \(\underline{73.8}\) / 73.6 & \(\underline{56.4}\) & 87.3 & \(\underline{86.4}\) & \(\bm{78.9}\) & 82.8 & \(\underline{51.6}\) & \(\bm{71.6}\)
& 51.6
& 326 & -- & \(\bm{85913}\) & 13433
& -- & -- & -- & --
& --
\\


\midrule
\multicolumn{19}{l}{
    \footnotesize
    \textsuperscript{\(\star\)} on CenterPoint~\cite{centerpoint} detections \hspace{5pt}
    \textsuperscript{\(\diamond\)} reproduced \hspace{5pt}
    \textsuperscript{\(\ddagger\)} deviating reported results
} \\
\bottomrule

\end{tabular}
}

\end{table*}

In the work at hand, we aim to find process and measurement noise covariance matrices that improve overall tracking performance.
We aim to maximize the \gls{amota} as a typical metric used in \gls{mot} evaluation.
Intuitively, the matrices $\bm{Q}$ and $\bm{R}$ should therefore realistically model the statistics of the dataset.
To isolate the effect of the noise modeling, we use a simple constant velocity model for the motion model and keep all other components of the tracking system fixed to the setup of Poly-MOT~\cite{poly_mot}. 
\subsection{Noise Model}
To optimize the noise estimate for a given dataset, we first need to define which parameters to optimize in the noise covariance matrices.
Let $c \in \mathcal{C}$ denote the semantic class of an object.
This work models class-dependent diagonal uncertainties
\[
\bm{Q}_c, \qquad \bm{R}_c.
\]
This means that noise in different elements of the state vector is assumed to be uncorrelated.
\subsubsection{Object-aligned Noise}
As the variances of the motion model noise in the longitudinal and lateral direction of vehicles in the street are not necessarily identical, we want the option to define covariance matrices that can describe this potential difference in acceleration behavior.
The purely diagonal definition of a noise covariance in world coordinates does not offer this option, so we formulate an object-aligned variant, where the principal axes of the ellipsoid described by this noise covariance are aligned with the object's frame axes $x^{\mathrm{obj}}$ and $y^{\mathrm{obj}}$.
This is achieved by rotating the covariance by a rotation matrix \(\bm{T}(\theta)\)
\begin{align}
\bm{Q}_{c}^{\mathrm{global}} &=
\bm{T}_{\mathrm{Q}}(\theta)\,
\bm{Q}_{c}^{\mathrm{obj}}\,
\bm{T}_{\mathrm{Q}}(\theta)^\top \label{eq:q_rotation}\\
\bm{R}_{c}^{\mathrm{global}} &=
\bm{T}_{\mathrm{R}}(\theta)\,
\bm{R}_{c}^{\mathrm{obj}}\,
\bm{T}_{\mathrm{R}}(\theta)^\top \label{eq:r_rotation}.
\end{align}
\(\bm{T}(\theta)\) differs for \(\bm{R}\) and \(\bm{Q}\) due to their different vector dimensions.
In both cases, it is constructed in such a way that it rotates an object frame position or velocity vector to the global frame.
The transformation matrix $\bm{T}_{\mathrm{Q}}(\theta)$ rotates planar position and velocity components $(x,y)$ and $(v_x,v_y)$ according to yaw $\theta$, while leaving $z$, $(w,l,h)$, $v_z$, and $\theta$ unchanged.
The transformation matrix $\bm{T}_{\mathrm{R}}(\theta)$ rotates planar measurement components $(x,y)$ and $(v_x,v_y)$ according to yaw $\theta$, while leaving $z$, $(w,l,h)$, and $\theta$ unchanged.
During application, the rotation matrices are constructed from the estimated yaw of the object, neglecting its uncertainty.

\subsubsection{Deriving Covariance Matrices Directly from Data}
To find a starting point for further investigation, we compute the sample covariance of the measurement and process noise for a constant velocity model of different object classes in both the local object and global map coordinate frames on the nuScenes dataset.

To compute the sample covariance of the measurement noise, first, detections of the CenterPoint~\cite{centerpoint} object detector are associated with the ground truth using the association strategy of the tracking framework \cite{poly_mot}.
Then, the residuals in the full measurement state between the detections and their matched ground truth are computed.
The sample covariance of these residuals is calculated to obtain an estimate of the measurement noise covariance matrices \(\bm{R}_c\).

The sample covariance of the process noise \(\bm{Q}_c\) is computed from the residuals of a one-step constant velocity motion prediction of ground truth tracks and their true future states.
All sample covariances are estimated using the unbiased estimator.

\subsection{Optimization}

Since \gls{amota} is obtained from the complete tracking and
evaluation pipeline and is not differentiable with respect to
the covariance parameters, we use gradient-free Bayesian
optimization~\cite{bayesian_optimization}.
The optimization is implemented using Weights \& Biases Sweeps ~\cite{wandb_sweeps} with \gls{amota} as the maximization objective.

To reduce computational cost, the class-aware optimization is
split into one subproblem per semantic class.
Each sweep optimizes the covariance parameters of one class using its class-wise \gls{amota}.
The first association stage is class-aware, resulting in negligible interactions between classes during tracking.

The diagonal covariance entries are optimized directly as positive variance parameters.
Each entry is searched within
\[
q_i,r_j \in [10^{-4},3]
\]
using a log-uniform search distribution.
The positive search interval ensures valid diagonal covariance matrices while covering several orders of magnitude.

We evaluate three optimized configurations.
CANMOT-SC optimizes the ten diagonal entries of the class-specific process noise covariance $\mathbf{Q}_c$, while fixing the measurement noise covariance $\mathbf{R}_c$ to the class-specific sample covariance estimated in the object frame over the trainval split.
CANMOT-Opt jointly optimizes the ten entries of $\mathbf{Q}_c$ and the nine entries of $\mathbf{R}_c$.
Shared-Opt optimizes the same 19 parameters, but uses one pair of covariance matrices shared across all classes.
CANMOT denotes the proposed method family comprising CANMOT-SC and CANMOT-Opt.

All optimization experiments use a budget of ten evaluations per optimized parameter. CANMOT-SC therefore uses 100 evaluations per class, whereas CANMOT-Opt uses 190 evaluations per class.
Shared-Opt uses 190 evaluations in total.
Each configuration is evaluated on the complete nuScenes validation set using fixed detector outputs and otherwise unchanged tracking parameters.
No early stopping is applied.
The configuration with the highest optimization objective is retained after completion of each sweep.

The same optimization protocol is applied to global and object-aligned noise models.
\section{Experimental Evaluation}
\label{sec:evaluation}
Evaluation is performed on the nuScenes dataset~\cite{nuscenes} using detections from CenterPoint~\cite{centerpoint}.
Following previous approaches, we optimize and evaluate the introduced method on the nuScenes val set to avoid using in-distribution data of the detector for tuning our tracking method.

\subsection{Metrics}
We report standard \gls{mot} evaluation metrics \gls{amota}, \gls{amotp}, \gls{ids}, \gls{frag}, \gls{tps}, and \gls{fps}.
In addition, we report the mean error of \gls{tps} in position \(e_\mathrm{t}\), size \(e_\mathrm{s}\), velocity \(e_\mathrm{v}\), and orientation \(e_\mathrm{o}\) as defined in \cite{nuscenes}.
Furthermore, we are interested in the consistency of our results, meaning whether the covariance estimated by the \gls{kf} matches the true error.
For this evaluation, we add the mean of the \gls{anees} over classes to the evaluation.
Since not all components of the ten-dimensional filter state are available in the ground-truth annotations, we evaluate ANEES in the nine-dimensional measurement space.
Mean \gls{anees} is calculated based on the squared Mahalanobis distance.
For \(C\) classes with \(N_c\) samples for class \(c\), we define 
\[
\mathrm{mANEES} = \frac{1}{C}\sum_{c=1}^{C}\frac{1}{N_c} \sum_{k=1}^{N_c}\tilde{\bm{z}_k}^\top \left(\hat{\bm{P}}^{\mathrm{z}}_k\right)^{-1}\tilde{\bm{z}_k},
\]
with \(\tilde{\bm{z}}_k = \bm{H}\hat{\bm{x}}_k - \bm{z}^{\mathrm{GT}}_k\) and \(\hat{\bm{P}}^{\mathrm{z}}_k = \bm{H}\hat{\bm{P}}_k\bm{H}^\top\).

\subsection{Discussion of Tracking Performance}
Table~\ref{tab:tracking_comparison} shows the described metrics for our method and several baselines grouped by local and global frame approaches.
The following observations are derived based on these results.

\textbf{R1: Local vs. Global Frame Modeling}.
\begin{figure*}[t]
    \centering
    \begin{subfigure}{0.44\textwidth}
        \includegraphics[width=\textwidth]{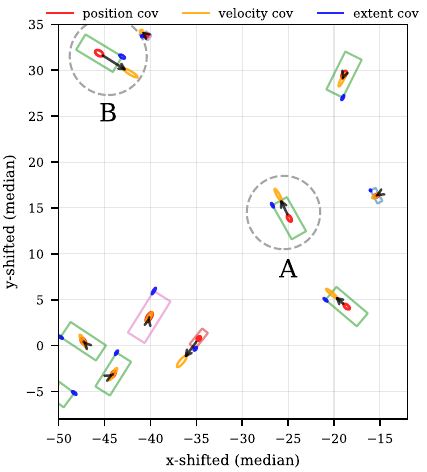}
        \caption{CANMOT-SC}
        \label{fig:qual_canmot}
    \end{subfigure}
    ~
    \begin{subfigure}{0.44\textwidth}
        \includegraphics[width=\textwidth]{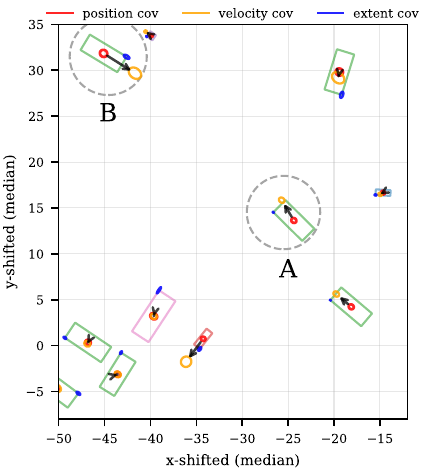}
        \caption{Global sample covariance (TV)}
        \label{fig:qual_gl_samp_cov}
    \end{subfigure}
    \\
    \begin{subfigure}{0.44\textwidth}
        \includegraphics[width=\textwidth]{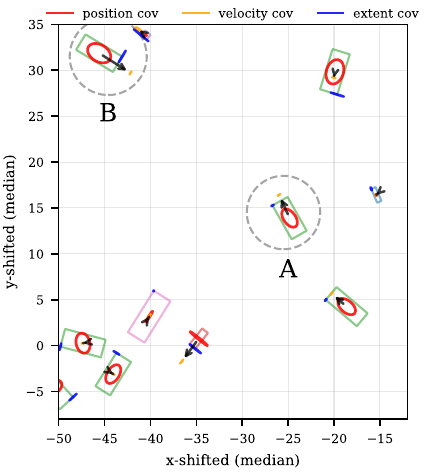}
        \caption{CANMOT-Opt}
        \label{fig:qual_canmot_sc}
    \end{subfigure}
    ~
    \begin{subfigure}{0.44\textwidth}
        \includegraphics[width=\textwidth]{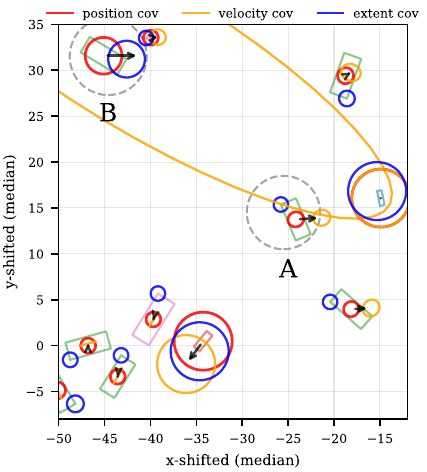}
        \caption{Poly-MOT \cite{poly_mot}}
        \label{fig:qual_poly_mot}
    \end{subfigure}
    \caption{
        Qualitative comparison of tracking results of (a) CANMOT-SC, (b) global sample covariance from trainval, (c) CANMOT-Opt, and (d) Poly-MOT for a sample scene from the nuScenes validation set.
        All objects are represented by their bounding box along with ellipses representing the estimated covariance of the tracks, including the uncertainty in position (red), velocity (yellow), and extent (blue).
        Bounding box colors indicate the class of the object, whereas arrows show the estimated velocity of the tracks.
        }
    \label{fig:qualitative_comparison}
\end{figure*}
The sample covariances (SC) computed on the training (-T), and trainval (-TV) sample set experience a slight degradation in terms of \gls{amota} (\(-\qty{0.1}{pp}\)) and \gls{amotp} (\(\approx+\qty{1}{m}\)) from global to local frame.
For the validation sample set (-V), an increase by \(\qty{0.4}{pp}\) in \gls{amota} and \(\qty{0.2}{m}\) in \gls{amotp} is observed.
This suggests a possible orientation bias in the validation set, due to the limited number of bicycle and motorcycle samples, and warrants further investigation.
The local frame sample covariances consistently outperform the global ones in terms of \gls{ids}, with the highest reduction of \(\qty{18.5}{\%}\) for the validation sample set.

The shared covariances optimized for the local frame outperform the global frame in terms of \gls{amota}, with an increase in \gls{ids} and \gls{frag}.
As \gls{ids} and \gls{frag} are reported as absolute numbers, they must be interpreted with respect to the number of true positives.
The local approach shows \(1076\) additional true positives.

Figure~\ref{fig:qualitative_comparison} shows an exemplary qualitative comparison of the tracking results for CANMOT-SC, the global sample covariance approach using trainval, CANMOT-Opt, and Poly-MOT~\cite{poly_mot}.
Here, the ellipses represent the estimated covariance of the tracks, including the uncertainty in position, size, and velocity.
The anisotropy of the covariance is clearly visible for CANMOT-SC, while the global sample covariance approach shows near isotropic covariances.
This leads to equal uncertainties in the lateral and longitudinal direction. \\
This is especially problematic for the car in the middle of the scene (A), where the orientation of the vehicle and the velocity are not well aligned.
The local approach shows a more realistic behavior of the vehicle turning right. 
Note that nuScenes \gls{amota} only depends on center distances. 
Thus, a better orientation and velocity estimation does not directly lead to a better \gls{amota}.
\\
Also, the near isotropic covariance of the velocity of the car in the upper left corner of the scene (B) leads to high uncertainty in the lateral direction, while the local approach shows a tighter covariance in the lateral direction, which is more realistic for a car driving straight.

CANMOT-Opt shows distinct behavior per state, where position estimates show larger estimated uncertainty, while those of velocity and size are smaller compared to the other approaches.

Figure~\ref{fig:qual_poly_mot} shows the results of Poly-MOT~\cite{poly_mot}.
The ellipses indicate unrealistically large estimated uncertainties in all states.
Such uncertainty estimates would provide limited utility for downstream modules and may lead to overly conservative behavior.
In addition, the estimated velocity directions are poorly aligned with the object headings for several vehicles, resulting in large apparent sideslip angles that are implausible under typical driving conditions.

The results indicate that locally modelled shared noise generalizes better over the different object classes.
Furthermore, local noise modeling shows a similar performance to global noise modeling in terms of \gls{amota}, but reduces \gls{ids} consistently.


\begin{table*}[t]
\centering
\caption{Uncertainty calibration analysis on nuScenes val with \gls{anees} and violation percentage \(p_\mathrm{v}\) of the \(\qty{95}{\%}\) confidence interval of \gls{nees} for different classes.}
\label{tab:calibration_comparison}

\setlength{\tabcolsep}{3pt}
\renewcommand{\arraystretch}{1.15}

\begin{tabular}{
l
*{16}{c}
}

\toprule

& \multicolumn{8}{c}{ANEES}
& \multicolumn{8}{c}{$p_\mathrm{v}$}
\\

\cmidrule(lr){2-9}
\cmidrule(lr){10-17}

Approach
& all & Bic. & Bus & Car & Motor. & Ped. & Tra. & Tru.
& all & Bic. & Bus & Car & Motor. & Ped. & Tra. & Tru.
\\

\midrule

\multicolumn{17}{c}{\cellcolor{gray!20}\textbf{Local}} \\
\midrule
CANMOT-SC
& {20.4}& {17.8} & {31.9} & {15.0} & {21.7} & {18.4} & {23.2} & {14.6}
& {24.6}& {38.9}& {45.1} & {20.4} & {39.1} & {29.4} & {39.5} & {25.1}
\\

CANMOT-Opt
& {2092.3} & {97.6}& {2985.7} & {7182.0} & {517.6} & {1965.5} & {666.9} & {1231.0}
& {93.8} & {82.9} & {99.7} & {93.0} & {94.1} & {94.5} & {96.4} & {96.7} 
\\

Shared-Opt
& {1902.5} & {2301.7} & {1305.8} & {1455.6} & {1838.7} & {1421.4} & {3714.8} & {1279.3}
& {96.1} & {96.1} & {99.3} & {95.1} & {96.2} & {97.4} & {99.4} & {98.0}
\\

\midrule

\multicolumn{17}{c}{\cellcolor{gray!20}\textbf{Global}} \\
\midrule

Poly-MOT\(\ddagger\)~\cite{poly_mot}
& {\(6\times10^{14}\)}& {1.4} & {\(2\times10^{15}\)} & {\(-5\times10^{15}\)} & {1.7} & {\(-1\times10^{16}\)} & {3576.9} & {\(2\times10^{16}\)}
& {47.1}& {0.1} & {54.8} & {52.2} & {0.6} & {36.4} & {94.6} & {52.1}
\\

Shared-Opt
& {7545.1}& {8767.2} & {6211.7} & {6660.6} & {9134.4} & {4947.7} & {12749.1} & {4345.0}
& {98.0}& {96.2} & {96.2} & {97.3} & {98.1} & {99.1} & {99.9} & {98.9}
\\

\midrule
\multicolumn{17}{l}{
    \footnotesize
    \textsuperscript{\(\ddagger\)} values not comparable as \(\widehat{P}\) lost positive semidefiniteness
} \\
\bottomrule

\end{tabular}

\end{table*}
\textbf{R2: Class-aware vs. Shared Noise Parameters}.
Optimizing the covariance matrices \(\bm{R}\) and \(\bm{Q}\) per class (CANMOT-Opt) yields the best \gls{amota}.
CANMOT-Opt improves \gls{amota} by \(\qty{0.5}{pp}\) over the shared local covariances with an equal \gls{amotp}.
\gls{ids} is reduced by \(\qty{8.0}{\%}\) and \gls{frag} by \(\qty{9.8}{\%}\).
CANMOT-Opt produces \(456\) fewer \gls{fps}, while providing \(210\) fewer \gls{tps}.
When optimizing \(\bm{Q}\) per class, while keeping \(\bm{R}\) from the sample covariance in the local frame (CANMOT-SC) the \gls{amota} decreases by \(\qty{0.3}{pp}\) compared to CANMOT-Opt.
This is still an increase of \(\qty{0.2}{pp}\) over the shared local covariances, while reducing \gls{ids} by \(\qty{15.9}{\%}\) and \gls{frag} by \(\qty{9.8}{\%}\).

Class-specific covariance parameterization improves \gls{amota} and reduces \gls{ids}.

\textbf{Comparison with \gls{sota}}.
As shown in Table~\ref{tab:tracking_comparison}, CANMOT-Opt achieves an \gls{amota} of \(\qty{73.8}{\%}\), matching IMM-MOT~\cite{IMM_MOT}.
It outperforms Poly-MOT~\cite{poly_mot} by \(\qty{1.0}{pp}\) and trails MCTrack~\cite{MCTrack} by \(\qty{0.2}{pp}\).
CANMOT-SC achieves a competitive \gls{amota} of \(\qty{73.5}{\%}\).
It exceeds Poly-MOT by \(\qty{0.7}{pp}\), trails IMM-MOT by \(\qty{0.3}{pp}\) to \(\qty{0.1}{pp}\), and remains \(\qty{0.5}{pp}\) below MCTrack.

Note that the advances of CANMOT are orthogonal to the advances of these methods, allowing further improvements when combining the approaches.

Poly-MOT produces the largest error in velocity estimation.
This stems from their high values in \(\bm{Q}\) and relatively low values in \(\bm{R}\) resulting in the \gls{kf} update replacing the state with the new measurement, instead of aggregating information over time.
These results suggest that improved motion modeling may enable tighter association gating thresholds. \\

CANMOT-SC provides more consistent uncertainty estimates as indicated by an \gls{anees} closer to the expected value of \(n_z=9\).
Tracks are also more consistent over time with \gls{ids} drastically reduced by \(\qty{19.1}{\%}\) over Poly-MOT~\cite{poly_mot}, \(\qty{41.7}{\%}\) over IMM-MOT~\cite{IMM_MOT}, and \(\qty{30.9}{\%}\) over MCTrack~\cite{MCTrack}.

\subsection{\bf{R3}: Consistency of Uncertainty Estimates}
\label{sec:results_uncertrainty}
For a consistent \gls{kf}, we expect \gls{anees} to be equal to the number of degrees of freedom \(n_z\) in the measurement space \cite{Chen2018weaknees}.
Large deviations of \gls{anees} from \(n_z\) indicate poor matching of the covariance to the true error, with large values indicating over-confidence, i.e. the filter is less accurate than estimated and smaller values indicating under-confidence, i.e. the filter is more accurate than estimated.

As Table~\ref{tab:tracking_comparison} shows, the sample-covariance variants yield moderate ANEES values of around \(40\) to \(50\).
as expected, SC-V performs best because it is estimated and evaluated on the validation set.
SC-T yields the highest ANEES, while SC-TV lies between SC-T and SC-V.
ANEES is largely independent of the reference frame.

Poly-MOT~\cite{poly_mot} provides uncertainty estimates that are not consistent with the true error, as indicated by the large \gls{anees}.
Please note that the estimated state covariances of Poly-MOT lost the property of being positive semi-definite in our experiments.
We attribute the loss of positive semidefiniteness to numerical instability induced by the unfavorable scaling of the released \(\bm{Q}\) and \(\bm{R}\) parameterization.
This led to negative \gls{anees} values for some classes, as seen in Table~\ref{tab:calibration_comparison}.
\(\widehat{\bm{P}}\) of Poly-MOT is thus not a valid covariance matrix, which breaks investigation of the consistency via \gls{anees}.

Using shared over individual covariances per class results in a lower \gls{anees}.
However, both approaches are not consistent with the true error, resulting in an \gls{anees} of \(2092.3\) for CANMOT-Opt and \(1902.5\) for Shared-Opt.

When optimizing \(\bm{Q}\) per class, while keeping \(\bm{R}\) from the sample covariance in the local frame (CANMOT-SC), the \gls{anees} is reduced to \(20.4\), which approaches the expected value of \(n_z=9\) for a consistent filter.

To test for filter consistency, \citeauthor{Chen2018weaknees}~\cite{Chen2018weaknees} propose counting violations of a defined confidence interval in the \(\chi^2\) distribution for the number of degrees of freedom of the filter.
Given, e.g. a \qty{95}{\percent} confidence interval, one expects the \gls{nees} to be inside this confidence interval for \qty{95}{\percent} of samples.
As over-confidence is more critical, we test for the upper threshold only, meaning that we treat each 
\[
\tilde{\bm{z}_k}^\top\left(\hat{\bm{P}}^{\mathrm{z}}_k\right)^{-1}\tilde{\bm{z}}_k > \chi^2_{n_z, 0.95}
\] as a violation.
Table~\ref{tab:calibration_comparison} shows the percentage of samples that are above the threshold as \(p_\mathrm{v}\) alongside the \gls{anees} for individual classes.
On this categorical distribution (i.e. below or above the threshold), we perform a \(\chi^2\)-test with a significance level of \(\qty{1}{\%}\) to see if for any of the evaluated methods, the number of samples that are inside the confidence interval differs significantly from the expected value.
Note that this can only be viewed as an explorative study due to limitations in the independence of the dataset used in our experiments.

Table~\ref{tab:calibration_comparison} shows that none of the evaluated \gls{kf}-based \gls{mot} methods provides consistent uncertainty estimates, and no violation distribution passes the \(\chi^2\)-test.
Methods with jointly optimized or manually tuned noise parameters \(\bm{Q}\) and \(\bm{R}\) exhibit violation probabilities \(p_\mathrm{v}>\qty{90}{\%}\) for almost all classes, with bicycles for CANMOT-Opt as the only exception.
This indicates that current tracking methods do not provide reliable uncertainty estimates for downstream tasks.
CANMOT-SC achieves the best consistency by optimizing \(\bm{Q}\) per class while retaining the local sample covariance for \(\bm{R}\).
It reaches an \gls{anees} of \(20.4\) and class-wise \(p_\mathrm{v}\) values between \(\qty{24.6}{\%}\) and \(\qty{45.1}{\%}\).
Although still well above the expected \(\qty{5}{\%}\), these results show that already subtle design changes can substantially improve filter consistency.

These preliminary results suggest that future work should incorporate calibration-aware objectives during covariance optimization.
\section{Conclusion and Outlook}
\label{sec:conclusion}
This work examined class-aware and object-aligned noise modeling in \gls{kf}-based \gls{mot} (CANMOT).
The study was guided by three research questions concerning coordinate frame selection, class-dependent modeling, and uncertainty consistency.

First, expressing process and measurement noise in the object coordinate frame preserves longitudinal-lateral anisotropy and consistently reduces identity switches, while maintaining comparable AMOTA to global formulations.
Second, class-specific covariance modeling improves tracking accuracy over shared noise parameters, indicating that heterogeneous object categories require distinct uncertainty representations.
Since CANMOT modifies the noise covariance parameterization rather than the filter structure itself, it remains compatible with any \gls{kf} variant.
Third, a calibration analysis based on \gls{anees} and \(\chi^2\)-test shows that widely used \gls{kf}-based \gls{mot} baselines are severely overconfident.
Optimizing the process noise per class while retaining sample-based measurement covariance substantially improves calibration, though full statistical consistency remains unattained.

These results highlight that uncertainty modeling constitutes a neglected design dimension in \gls{kf}-based \gls{mot}.
Future work will investigate structured covariance parameterizations beyond diagonal assumptions, consistency-aware optimization criteria, and integration with multiple motion models.

\printbibliography



\end{document}